\documentclass[conference]{IEEEtran}
\IEEEoverridecommandlockouts
\usepackage{cite}
\usepackage{amsmath,amssymb,amsfonts}
\usepackage{algorithmic}
\usepackage{graphicx}
\usepackage{textcomp}
\usepackage{hyperref}
\usepackage{xcolor}

\def\BibTeX{{\rm B\kern-.05em{\sc i\kern-.025em b}\kern-.08em
    T\kern-.1667em\lower.7ex\hbox{E}\kern-.125emX}}
\usepackage{tikz}
\usepackage{textcomp}
\usepackage{hyperref}
\usepackage{lipsum}

\newcommand\copyrighttext{%
  \footnotesize \textcopyright 2025 IEEE. Personal use of this material is permitted.
  Permission from IEEE must be obtained for all other uses, in any current or future
  media, including reprinting/republishing this material for advertising or promotional
  purposes, creating new collective works, for resale or redistribution to servers or
  lists, or reuse of any copyrighted component of this work in other works.}
\newcommand\copyrightnotice{%
\begin{tikzpicture}[remember picture,overlay]
\node[anchor=south,yshift=10pt] at (current page.south) {\fbox{\parbox{\dimexpr\textwidth-\fboxsep-\fboxrule\relax}{\copyrighttext}}};
\end{tikzpicture}%
}

\begin{document}

\title{Explore Activation Sparsity in Recurrent LLMs for Energy-Efficient Neuromorphic Computing}

\author{\IEEEauthorblockN{Ivan Knunyants\IEEEauthorrefmark{1}, Maryam	Tavakol\IEEEauthorrefmark{2}, Manolis	Sifalakis\IEEEauthorrefmark{1}, Yingfu	Xu\IEEEauthorrefmark{1}, Amirreza	Yousefzadeh\IEEEauthorrefmark{1}\IEEEauthorrefmark{3}, Guangzhi Tang\IEEEauthorrefmark{4}}
\IEEEauthorblockA{\textit{\IEEEauthorrefmark{1}imec Netherlands, Eindhoven, Netherlands}, \textit{\IEEEauthorrefmark{2}TU Eindhoven, Eindhoven, Netherlands}} \IEEEauthorblockA{\textit{\IEEEauthorrefmark{3}University of Twente, Enschede, Netherlands}, \textit{\IEEEauthorrefmark{4}Maastricht University, Maastricht, Netherlands}}
}

\maketitle
\copyrightnotice

\begin{abstract}
The recent rise of Large Language Models (LLMs) has revolutionized the deep learning field. However, the desire to deploy LLMs on edge devices introduces energy efficiency and latency challenges. Recurrent LLM (R-LLM) architectures have proven effective in mitigating the quadratic complexity of self-attention, making them a potential paradigm for computing on-edge neuromorphic processors. In this work, we propose a low-cost, training-free algorithm to sparsify R-LLMs' activations to enhance energy efficiency on neuromorphic hardware. Our approach capitalizes on the inherent structure of these models, rendering them well-suited for energy-constrained environments. Although primarily designed for R-LLMs, this method can be generalized to other LLM architectures, such as transformers, as demonstrated on the OPT model, achieving comparable sparsity and efficiency improvements. Empirical studies illustrate that our method significantly reduces computational demands while maintaining competitive accuracy across multiple zero-shot learning benchmarks. Additionally, hardware simulations with the SENECA neuromorphic processor underscore notable energy savings and latency improvements. These results pave the way for low-power, real-time neuromorphic deployment of LLMs and demonstrate the feasibility of training-free on-chip adaptation using activation sparsity.
\end{abstract}


\section{Introduction}

Large language models (LLMs), such as GPT-4\cite{achiam2023gpt}, and open-source models, like LLaMA-3\cite{dubey2024llama}, have become pivotal in advancing natural language processing. Their ability to process and generate high-quality text has led to widespread adoption across various industries. However, as LLMs and training datasets become larger, they demand more computational resources for training and inference. Specifically, the self-attention mechanism of LLM architectures scales quadratically with the input length, while the computations in linear layers also increase quadratically with the growing dimension of tokens. These computational complexities are even more problematic when deploying LLMs on Edge AI devices with limited computing and memory resources. Besides, data-specific optimizations are often too resource-intensive to execute directly on the device, leading to privacy concerns when transferring data between devices and servers.

Recurrent LLMs (R-LLMs), such as Transformer-RNN fusion models RWKV\cite{Peng2023RWKVRR},  RetNet\cite{Sun2023RetentiveNA}, xLSTM\cite{Beck2024xLSTMEL}, and Griffin\cite{De2024GriffinMG}, as well as state space models like S4\cite{Gu2021EfficientlyML} and Mamba\cite{Gu2023MambaLS,Dao2024TransformersAS}, have emerged as lighter alternatives to self-attention LLMs. These methods combine the ability of recurrent inference and operate with linear complexity, addressing the quadratic self-attention issue while still benefiting from the rapid parallel training. Additionally, the recurrent computational paradigm is well-suited for neuromorphic processors, which are brain-inspired low-power AI-dedicated hardware\cite{tang2023seneca,Modha2023IBMNN,Orchard2021EfficientNS}. These processors manage event-based data-flow processing, exploiting the activation sparsity in neural networks for energy-efficient and low-latency computation. To maximize the benefits of neuromorphic computing, a high level of activation sparsity in the neural networks is crucial \cite{Xu2024OptimizingEN}. However, R-LLM models consist of dense linear layers, including high-dimensional upward and downward projections, resulting in costly computation on neuromorphic processors. Therefore, there is a need to explore activation sparsity in R-LLMs for energy-efficient neuromorphic computing.

State-of-the-art methods for improving activation sparsity in LLMs apply the ReLU activation function before linear layers in pre-trained models and perform fine-tuning on large datasets to restore performance\cite{Mirzadeh2023ReLUSB,Song2024ProSparseIA}. Moreover,\cite{Song2024ProSparseIA} applied activation regularization in loss functions to enable sparse-aware fine-tuning and further increased sparsity. However, these approaches involve a training-based fine-tuning stage, which requires high computational costs due to the model size and training tokens. Privacy concerns over local fine-tuning data make on-device LLM adaptations preferable\cite{li2024llm}, but training computational costs render this impractical. Furthermore, current solutions are designed for transformer-based models, and to our knowledge, no low-cost activation sparsification method exists for R-LLMs that are also well-suited for on-device adaptation.

This paper proposes an R-LLM with activation sparsity for energy-efficient neuromorphic computing and a training-free threshold adaptation algorithm for improving activation sparsity.\footnote{Code: https://github.com/ERNIS-LAB/LLM-activation-sparsity} Our contributions are as follows:
\begin{itemize}
    \item We introduce an event-based R-LLM with thresholding functions. The resulting optimized R-LLM obtains an average activation sparsity up to 63\%, increasing $2.2\times$ compared to the original model with natural sparsity\cite{Peng2023RWKVRR}.
    \item We propose a training-free algorithm to find thresholds using local data adaptively. The algorithm can be deployed on neuromorphic processors and is $30\times$ more efficient on GPU than the training-based method\cite{Mirzadeh2023ReLUSB}.
    \item We demonstrate a $1.9\times$ energy and latency improvement of the sparse model via hardware simulation with the SENECA neuromorphic processor\cite{tang2023seneca}.
    \item We extend our approach to a self-attention LLM (OPT\cite{zhang2022opt}), demonstrating comparable results to the training-based method\cite{Mirzadeh2023ReLUSB}.
\end{itemize}

\section{Neuromorphic Recurrent LLMs}

\subsection{Activation Sparsity in Recurrent LLMs}

We present an efficient approach to sparsify the activation of recurrent LLMs (R-LLMs), inspired by neuromorphic computing principles, that can be trained on a small dataset without requiring costly fine-tuning. The core idea is integrating activation sparsity before the linear layers, automatically zeroing out insignificant values during matrix multiplications. Following the concept of weight sparsity, we assume that activations with small absolute values are unimportant, as they have minimal impact on computations. In our approach, we use a thresholding function with a parameter $\lambda$, setting values with an absolute magnitude lower than $\lambda$ to zero, 

$$ \text{Threshold}(x, \lambda) = \begin{cases} 
x & \text{if } |x| \geq \lambda, \\
0 & \text{otherwise}.
\end{cases}$$

To this end, we iterate over all linear layers and initialize the sparsifying function before each layer by performing a training-free threshold search using a small dataset. Given the varying importance of different layers of the architecture, their respective activation functions are set to different sparsity levels to optimize the overall average sparsity. As a result, we place these sparsifying thresholding functions before all linear layers of the targeted R-LLM for energy-efficient neuromorphic processing.

\begin{figure}[h!]
    \centering
\includegraphics[width=0.45\linewidth]{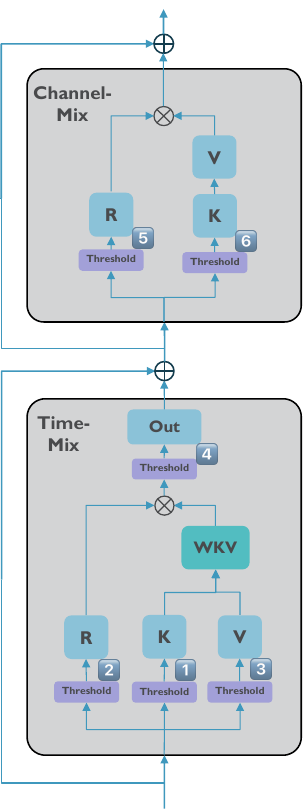}
    \caption{Recurrent LLM (RWKV) block with proposed thresholding function for activation sparsity. The thresholds are initialized in each of the six thresholding functions following the presented order from 1 to 6 in the figure.}
    \label{fig:rwkv_block_sparse}
\end{figure}

We use RWKV\cite{Peng2023RWKVRR} as the R-LLM representative to demonstrate the effectiveness of our proposed method. The RWKV consists of Time-Mix and Channel-Mix sub-blocks. A Time-Mix sub-block with linear recurrent processing offers an alternative to the self-attention mechanism. This block applies three linear projections: \textbf{R}, \textbf{K} and \textbf{V}, to the tokens and mixes them with the previous tokens in the WKV recurrent mechanism. The \textbf{Out} linear projection precedes the addition of the residual connection. The Channel-Mix uses a receptance linear layer (\textbf{R}) while also using the classic LLM structure of upward (\textbf{K}) and downward (\textbf{V}) projections to/from $4\times$ dimension of embeddings. Inputs to the six linear layers are sparsified with the proposed thresholding function (see Figure~\ref{fig:rwkv_block_sparse}), while the downward projection (\textbf{V} in Time-Mix) has a natural sparse input due to the $\text{ReLU}^2$ function before the linear layer in the original structure of RWKV.

\subsection{Efficient Training-free Threshold Initialization Algorithm}

\begin{figure}[h!]
    \centering
    \includegraphics[width=1\linewidth]{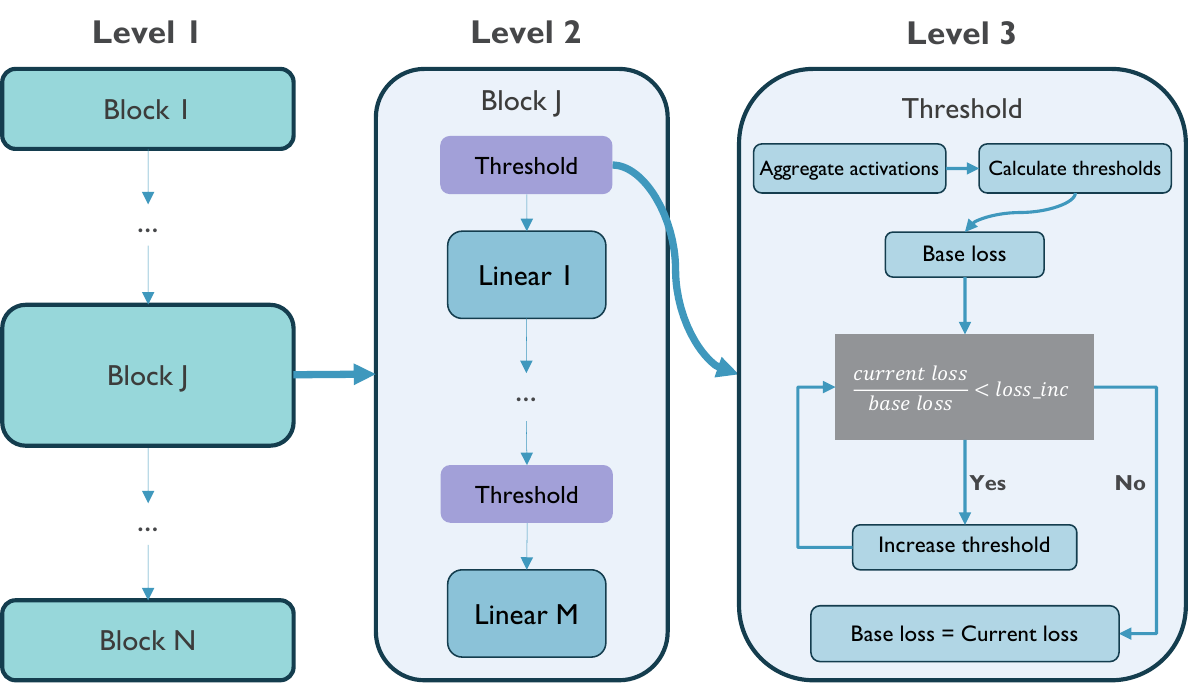}
    \caption{Threshold initialization algorithm. \textbf{Level 1:} iterate over LLM blocks. \textbf{Level 2:} in each block iterate over thresholding functions following the pre-defined order. \textbf{Level 3:} search the optimal threshold for each thresholding function by performing R-LLM inference. If we have $k_{num}$ possible \textit{sparsity percentages}, the algorithm will iterate over $N \times M$ threshold functions, with each function being initialized $k_{num}$ times in the worst-case scenario.}
    \label{fig:topk_iter_algo}
\end{figure}

We propose a training-free algorithm to efficiently initialize the thresholding functions in pre-trained R-LLMs (see Figure \ref{fig:topk_iter_algo}). Since each initialization step modifies the activation distribution of subsequent layers, the algorithm iterates activations/blocks sequentially from the first to the last layer/block. Since all blocks in the R-LLM share the same structure, the order of initializing the thresholding functions within each block remains consistent. Figure \ref{fig:rwkv_block_sparse} exemplifies the initialization order within RWKV blocks, with components numbered.

The thresholding functions must be initialized with a certain sparsity level, ensuring a controllable trade-off between sparsity and model performance. Before each threshold initialization, we record the activation values on a small initialization dataset. We use these values to determine the thresholds corresponding to the average \textit{sparsity percentages} we want to consider. For example, we can calculate threshold values that provide us with 20\%, 60\%, and 80\% sparsity on the initialization dataset.

We aim to identify the highest possible \textit{sparsity percentages} for the thresholding function while maintaining the model's performance. We assess the model's performance by evaluating the loss on the initialization dataset. Initially, we have a \textit{base loss} representing the loss value before applying the thresholding function. Then, for the first R-LLM block, we iterate over increasing \textit{sparsity percentages'} thresholds, while the condition $\frac{\textit{current loss}}{\textit{base loss}} < loss\_inc$ holds. 
This continues until the loss increases significantly based on the parameter $loss\_inc$. For the following blocks, the algorithm cycles through \textit{sparsity percentages}, beginning with the most commonly used percentage in the previous blocks at the same thresholding function position. This is due to our assumption that the optimal \textit{sparsity percentages} are similar between the identically structured R-LLM blocks.

The proposed threshold initialization algorithm is training-free and only requires iterative inferences of R-LLMs with sparse activation. Therefore, it can be applied to neuromorphic processors for energy-efficient adaptation of the thresholds using locally collected data.

\section{Empirical Study and Results}

\subsection{Experimental Setups}

To iteratively compute the loss during the threshold initialization algorithm, we utilize the Minipile dataset~\cite{Kaddour2023minipile}, which features diverse data intersecting with the RWKV's pre-trained dataset. Minipile is a 6GB filtered and deduplicated subset of the Pile dataset~\cite{Gao2020pile} for pre-training the RWKV. In our experiments, we randomly sample 1k documents from Minipile's 1M training set as the initialization dataset and utilize a separate test dataset of 10k documents for final loss and sparsity calculation. We iterate over linearly increasing \textit{sparsity percentages} $\in [10, 20, ..., 80, 90]$ for thresholding functions, and use $loss\_inc = 1.0005$.

\subsection{Minipile Test Performance Results}

We first evaluate the performance of our proposed approach on three different-sized pre-trained RWKV models (with 430M, 1.5B, and 3B parameters). We measure our models' sparsity and loss on the Minipile test dataset to obtain reliable values and compare against the original dense models with natural sparsity (baseline) on Table~\ref{table:spars_loss_rwkv}. On average, the original models show a low sparsity level of around 28\% due to the natural sparsity but exhibit $\approx5\%$ lower test loss compared to our models. For instance, for the smaller model with 430M parameters, the baseline shows a test loss of 2.2377, whereas the proposed approach, with a significantly higher sparsity of 57.03\%, results in a slightly increased test loss of 2.3377, indicating a 4.47\% increase in loss. This pattern holds across the larger models as well. The 3B baseline model demonstrates a test loss of 1.9297, whereas the proposed approach results in a higher test loss of 2.0510, corresponding to a 6.29\% increase while achieving a 2.2x increase in sparsity. Despite the slight increase in the loss, the higher sparsity achieved by the proposed approach leads to more efficient computation and memory savings. Therefore, balancing accuracy with model efficiency is a crucial consideration for large-scale deployment.

\begin{table}[t]
\centering
\caption{Sparsity vs. test loss in Various RWKV Model Sizes}
\scalebox{0.9}{
\begin{tabular}{|c|l|c|c|c|}
\hline
\textbf{Model size} & \textbf{Model type} & \textbf{Sparsity (\%)} & \textbf{Test loss} & \textbf{Loss Increase (\%)} \\
\hline
430M & Baseline\cite{Peng2023RWKVRR} & 28.01 & 2.2377 &  \\
& Our approach & 57.03 & 2.3377 & 4.47 \\
\hline
1.5B & Baseline\cite{Peng2023RWKVRR} & 28.38 & 2.0222 & \\
 & Our approach & 59.99 & 2.1111 & 4.40 \\
\hline
3B & Baseline\cite{Peng2023RWKVRR} & 28.65 & 1.9297 & \\
 & Our approach & 63.16 & 2.0510 & 6.29 \\
\hline
\end{tabular}
}
\label{table:spars_loss_rwkv}
\end{table}

\begin{table*}[h!]
\centering
\caption{Zero-Shot Benchmark Results for RWKV Models with and without proposed thresholding algorithm}
\scalebox{0.8}{
\begin{tabular}{|c|c|c|c|c|c|c|c|c|c|c|}
\hline
\textbf{Model size} & \textbf{Model type} & \textbf{AVG} & \textbf{PIQA} & \textbf{WinoGrande} & \textbf{HellaSwag} & \textbf{ARC-E} & \textbf{ARC-C} & \textbf{LAMBADA} & \textbf{OpenBookQA} & \textbf{SciQ}\\
\hline
430M & Baseline\cite{Peng2023RWKVRR}     & 49.4 & 67.8 & 51.9 & 40.7 & 52.7 & 25.0 & 46.0 & 31.6 & 79.2 \\
& Our approach & 47.6 & 65.7 & 50.7 & 39.4 & 51.6 & 25.8 & 37.1 & 31.0 & 79.9 \\
\hline
1.5B & Baseline\cite{Peng2023RWKVRR}     & 55.6 & 72.2 & 54.5 & 52.8 & 60.6 & 29.2 & 57.4 & 34.0 & 84.1 \\
& Our approach & 53.1 & 70.8 & 52.9 & 49.9 & 58.4 & 28.0 & 49.0 & 31.8 & 83.7 \\
\hline
3B   & Baseline\cite{Peng2023RWKVRR}     & 59.6 & 73.7 & 60.2 & 59.6 & 64.5 & 31.8 & 64.0 & 37.2 & 85.5 \\
& Our approach & 55.5 & 70.7 & 56.0 & 55.5 & 61.2 & 30.5 & 52.1 & 35.2 & 83.0 \\
\hline
\end{tabular}
}\label{tab:bench}
\end{table*}

\subsection{Performance on Zero-Shot Learning Benchmarks}

As the loss values do not accurately reflect the degradation in language modeling capabilities, we measure the performance of our models on multiple well-known LM benchmarks in the Language Model Evaluation Harness\cite{gao2021framework}, PIQA, Winogrande, Hellaswag, ARC Challenge, ARC Easy, LAMBADA, OpenBookQA, and SciQ. Table~\ref{tab:bench} outlines the obtained results from this analysis.

The benchmark results demonstrate that our approach maintains a robust overall performance across a wide range of tasks while significantly increasing sparsity. While there is a slight reduction in accuracy compared to the baseline models, the drop is minimal, indicating that our approach effectively balances sparsity and efficiency with only a minor impact on performance. This consistent pattern across various benchmarks highlights the robustness of our approach, making it a compelling solution for optimizing resource usage while still achieving competitive results.

\subsection{Effectiveness of Sparsity Percentage Search}

In the threshold initialization algorithm, we use the fact that all R-LLM blocks share the same structure. We can anticipate similar behavior across corresponding thresholding function positions in each block. Therefore, we used a heuristic to start the iteration process with the most commonly used percentage in the previous blocks at the same position. If we do not exploit this heuristic, we will iterate over all \textit{sparsity percentages}, starting with the lowest. Therefore, without the heuristic, the initialization algorithm will result in an increase in runtime (number of inferences for initialization) by around 3 times in our experiment.

\section{Hardware Simulation}

\subsection{Simulation with SENECA Neuromorphic Processor}

To clearly understand the benefit of activation sparsity improvements, we conduct an analytical hardware simulation study to evaluate the key hardware metrics: energy and latency. The study is performed using the SENECA neuromorphic processor\cite{tang2023seneca} and based on actual hardware measurements in\cite{Tang2023OpenTB}. Each input token processing in RWKV uses the same number of dense computations. Our energy and latency studies focus on processing a single token within one RWKV block and use sparsity values averaged over RWKV blocks and tokens within the test dataset. The results can be multiplied by the number of layers and tokens to estimate the performance of an entire R-LLM and document. We compare the baseline (Dense) with the sparsified (Sparse) version of the 3B RWKV.

\subsection{Hardware Simulation Results}

We present the results of two parts of the RWKV model: Time-Mix and Channel-Mix sub-blocks. The comparisons for energy and latency are very similar, as both depend linearly on the total operation count. We focus on two RWKV sub-blocks and omit LayerNorm due to its negligible impact. The hardware operations consist of memory operations (read/write) and computational operations (add, mult, div, etc.). Even though the RWKV block uses functions like sigmoid, token shift, and others, around 97\% of all operations are made by linear layers. Table~\ref{tab:energy} and \ref{tab:latency} respectively illustrate the results for energy and latency. We can observe that the influence of sparsification is considerable for Time-Mix by $2.3\times$ because of the highly sparsified R, V, K, and Out linear layers. However, Channel-Mix is improved by $1.6\times$ as we counted the natural sparsity of the original RWKV model, and the only gain is in R and K linear layers. Nevertheless, both latency and energy metrics in the entire model are improved by $1.9\times$, which is a considerable change, while the sparsified model still shows competitive language modeling abilities.

\begin{table}[t]
\centering
\caption{Energy ($\mu J$) cost on SENECA}
\begin{tabular}{|c|c|c|c|c|c|c|}
\hline
 & \multicolumn{2}{c|}{Time-Mix} & \multicolumn{2}{c|}{Channel-Mix} & \multicolumn{2}{c|}{Overall} \\ \cline{2-7} 
                  & Sparse        & Dense         & Sparse          & Dense         & Sparse        & Dense         \\ \hline
Computation       & 5.0           & 11.9          & 9.3             & 15.5          & 14.3          & 27.4          \\ \hline
Memory            & 7.4           & 17.6          & 13.9            & 23.1          & 21.3          & 40.7          \\ \hline
Total             & \textbf{12.4} & 29.5 & \textbf{23.2}   & 38.6 & \textbf{35.6} & 68.1 \\ \hline
\end{tabular}\label{tab:energy}
\end{table}

\begin{table}[h!]
\centering
\caption{Latency ($ms$) on SENECA}
\begin{tabular}{|c|c|c|c|c|c|c|}
\hline
 & \multicolumn{2}{c|}{Time-Mix} & \multicolumn{2}{c|}{Channel-Mix} & \multicolumn{2}{c|}{Overall} \\ \cline{2-7} 
                  & Sparse        & Dense         & Sparse          & Dense         & Sparse        & Dense         \\ \hline
Computation       & 0.9           & 2.1           & 1.7             & 2.8           & 2.6           & 4.9           \\ \hline
Memory            & 1.3           & 3.1           & 2.5             & 4.1           & 3.8           & 7.2           \\ \hline
Total             & \textbf{2.2}  & 5.2  & \textbf{4.2}    & 6.9  & \textbf{6.4}  & 12.1 \\ \hline
\end{tabular}\label{tab:latency}
\end{table}

\section{Generalization to Self-Attention LLM}

Our approach can be extended to the standard transformer architectures. Tables \ref{tab:opt_sparse} and \ref{tab:opt_bench} compare the application of our method on a 2.7B pre-trained OPT model\cite{zhang2022opt} with the state-of-the-art fine-tuning-based approach that requires costly training\cite{Mirzadeh2023ReLUSB}. The results indicate that our approach performs on par with \cite{Mirzadeh2023ReLUSB} with similar activation sparsity and zero-shot benchmark accuracy. Moreover, we show results with different sparsity and accuracy tradeoffs by controlling the loss\_inc parameter in the thresholding initialization algorithm. Specifically, when loss\_inc=1.0004, the model shows slightly better overall sparsity and a slight accuracy increase in average zero-shot learning benchmarks.

\begin{table}[h!]
\centering
\caption{Sparsity and Averaged Zero-Shot Accuracy of OPT Models}
\scalebox{0.7}{
\begin{tabular}{|c|ccc|c|c|}
\hline
\textbf{Model} & \multicolumn{3}{c|}{\textbf{Activation sparsity (\%)}} & \textbf{Overall sparsity} & \textbf{AVG Benchmark} \\
& \textbf{QKV} & \textbf{UpProj} & \textbf{DownProj} & &  \textbf{Accuracy (\%)}\\ 
\hline
2.7B Base\cite{zhang2022opt} & 0 & 0 & 96 & 48 & 60.3 \\
2.7B Training-based\cite{Mirzadeh2023ReLUSB} & 50 & 35 & 96 & 71.125 & 58.5 \\
2.7B Our (loss\_inc = 1.0003) & 46 & 35 & 97 & 70.125 & 59.8 \\
2.7B Our (loss\_inc = 1.0004) & 48 & 38 & 97 & 71.25 & 58.6 \\
2.7B Our (loss\_inc = 1.0005) & 50 & 39 & 97 & 72.125 & 58.3 \\
\hline
\end{tabular}
}\label{tab:opt_sparse}
\end{table}

\begin{table}[h!]
\centering

\caption{Zero-Shot Accuracy (\%) of the OPT Model on LM Benchmarks}
\scalebox{0.65}{
\begin{tabular}{|c|ccccccc|}
\hline
\textbf{Model} & \textbf{PIQA} & \textbf{WinoGrande} & \textbf{Hellaswag} & \textbf{Arc-E} & \textbf{Arc-C} & \textbf{LAMBADA} & \textbf{SciQ} \\
\hline
2.7B Base\cite{zhang2022opt} & 73.8 & 61.0 & 45.8 & 60.8 & 31.2 & 63.6 & 85.8 \\
2.7B Training-based\cite{Mirzadeh2023ReLUSB} & 73.9 & 59.6 & 44.9 & 60.3 & 26.8 & 57.6 & 86.7 \\
2.7B Our (loss\_inc = 1.0003) & 72.8 & 61.8 & 43 & 57.3 & 30.1 & 66.8 & 87.1 \\
2.7B Our (loss\_inc = 1.0004) & 71.5 & 61 & 42.4 & 56.4 & 30.1 & 63.3 & 85.6 \\
2.7B Our (loss\_inc = 1.0005) & 72.1 & 60.6 & 42.1 & 56.3 & 29.4 & 63.1 & 84.8 \\
\hline
\end{tabular}
}\label{tab:opt_bench}
\end{table}

Overall, our training-free algorithm performs better or on par compared to the costly state-of-the-art training-based method\cite{Mirzadeh2023ReLUSB}. It is important to emphasize the main benefit of our algorithm, where its overall computation cost is significantly lower than training-based fine-tuning. Regarding GPU hours, we estimate our approach to be $30\times$ more efficient than fine-tuning the same model on 50B tokens. Since our proposed method only performs network inference, it will also consume less memory than the training-based methods, resulting in less area needed for the hardware. The advantage will become even more significant when we consider applying our training-free threshold initialization algorithm on dedicated hardware that can exploit the activation sparsity in self-attention LLMs.

\section{Conclusion}

This paper introduced an efficient activation sparsification technique for recurrent LLMs (R-LLMs). Our method significantly reduces computation and energy demands without using training-based fine-tuning, making it ideal for resource-constrained applications. Experiments show that our approach delivers competitive performance with high activation sparsity, while hardware simulations on the SENECA neuromorphic processor demonstrate substantial energy efficiency and latency gains. Consequently, this work advances the deployment of dedicated LLMs for neuromorphic computing, facilitating low-power, real-time generative AI applications.

\section*{\small{Acknowledgment}}
\small{This work used the Dutch national e-infrastructure with the support of the
SURF Cooperative using grant no. EINF-10787.}

\newpage
\bibliographystyle{IEEEtran}
\bibliography{ref}
\end{document}